\pdfoutput=1

\documentclass[11pt]{article}

\usepackage[]{EMNLP2023}

\usepackage{times}
\usepackage{latexsym}

\usepackage[T1]{fontenc}

\usepackage[utf8]{inputenc}

\usepackage{microtype}

\usepackage{inconsolata}

\usepackage{tikz}
\usetikzlibrary{shapes, arrows.meta, positioning}

\usepackage{mathrsfs}
\usepackage{amsmath}
\usepackage{graphicx}
\usepackage{float}
\usepackage{booktabs}
\usepackage{colortbl}
\usepackage{tabularx}
\usepackage{pbox}
\usepackage{amsfonts, amssymb}
\usepackage{cleveref}
\usepackage{hyperref}
\usepackage{makecell}
\usepackage{verbatim}
\usepackage{paralist}
\usepackage[usestackEOL]{stackengine}
\PassOptionsToPackage{table}{xcolor}
\usepackage{mathtools, nccmath}

\usepackage{lipsum}

\usepackage{enumitem}

\usepackage[ruled,vlined]{algorithm2e}
\usepackage{algpseudocode}

\usepackage{caption}
\usepackage{subcaption}

\usepackage{pifont}
\usepackage{arydshln}

\usepackage{booktabs}

\usepackage{hyperref}

\title{Pit One Against Many: Leveraging Attention-head Embeddings for Parameter-efficient Multi-head Attention}

\author{Huiyin Xue \and Nikolaos Aletras \\
        Department of Computer Science, University of Sheffield \\
United Kingdom \\
 \texttt{\{hxue12, n.aletras\}@sheffield.ac.uk}}

\begin{document}
\maketitle

\begin{abstract}

Scaling pre-trained language models has resulted in large performance gains in various natural language processing tasks but comes with a large cost in memory requirements. Inspired by the position embeddings in transformers, we aim to simplify and reduce the memory footprint of the multi-head attention (MHA) mechanism. We propose an alternative module that uses only a single shared projection matrix and multiple head embeddings (MHE), i.e. one per head. We empirically demonstrate that our MHE attention is substantially more memory efficient compared to alternative attention mechanisms while achieving high predictive performance retention ratio to vanilla MHA on several downstream tasks. MHE attention only requires a negligible fraction of additional parameters ($3nd$, where $n$ is the number of attention heads and $d$ the size of the head embeddings) compared to a single-head attention, while MHA requires $(3n^2-3n)d^2-3nd$ additional parameters.\footnote{Code: \url{https://github.com/HUIYINXUE/simpleMHE}}

\end{abstract}

\section{Introduction}

Scaling pre-trained language models (PLMs) aims to enhance performance by increasing their size and capacity, leading to models with an unprecedented number of parameters~\citep{kaplan2020scaling,chowdhery2022palm,hoffmann2022training}. Just by increasing the size of PLMs and the pre-training data has yielded state-of-the-art performance on various natural language processing (NLP) tasks~\citep{devlin-etal-2019-bert,liu2019roberta,clark2020electra,raffel2020exploring,brown2020language,clark-et-al2022-canine,ouyang2022training,touvron2023llama}.

However, the pursuit of developing larger PLMs comes with large computational requirements. This has direct environmental implications such as large carbon emissions~\citep{lacoste2019quantifying,strubell-etal-2019-energy,weidinger2022taxonomy}, conflicting with the principles of Green artificial intelligence development~\citep{schwartz2020green}. Moreover, scaling can hinder researchers with limited access to computing resources to participate in advancing the field~\citep{schwartz2020green}. This results in inequalities, where only a privileged few can actively contribute, potentially impeding diversity and inclusivity~\citep{weidinger2022taxonomy}.

\begin{figure}[!t]
    \centering
    \includegraphics[width=0.97\linewidth]{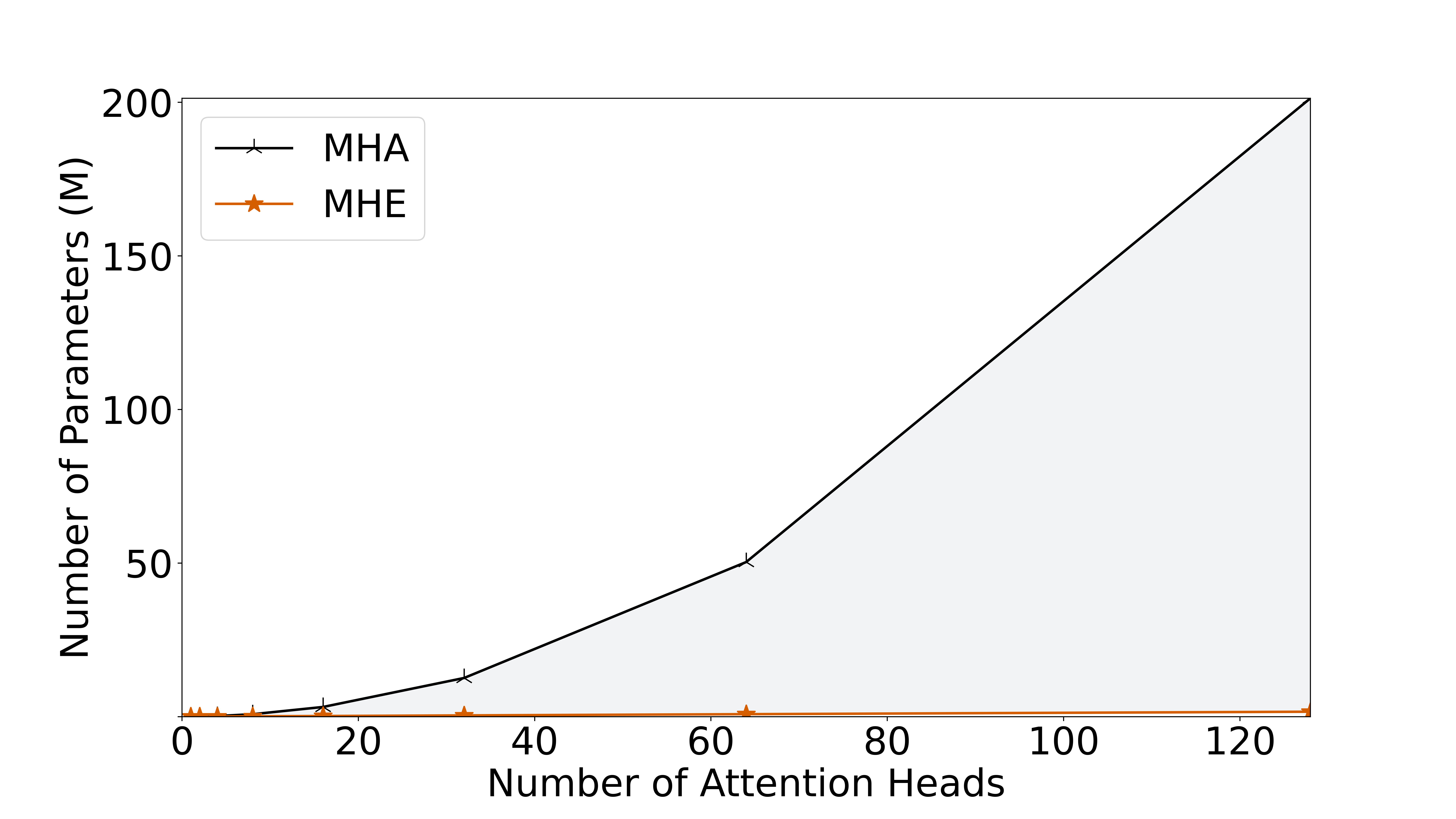}
    \caption{Number of parameters for an attention sublayer and different number of attention heads using multi-head attention \textsc{MHA} and our multi-head embedding attention \textsc{MHE}. We fix the dimension of attention to 64, only counting the parameters for projecting queries, keys, and values.}
    \label{fig:scaling_params_intro}
\end{figure}

The backbone of transformers~\citep{vaswani2017attention} is the multi-head attention (\textsc{MHA}) module that extends the standard single-head attention (\textsc{SHA}) proposed by \citet{cho-etal-2014-properties}. \textsc{MHA} applies an attention mechanism (i.e. head) multiple times for the same set of queries, keys and values by using a different set of parameters (i.e. projection matrices) for each of them. This results in \textsc{MHA} modules with a large memory footprint that increases with the number of layers and attention heads per layer in PLMs~\citep{devlin-etal-2019-bert,brown2020language,ouyang2022training,touvron2023llama}. Figure \ref{fig:scaling_params_intro} shows how the number of parameters of a single attention sublayer increases with its number of attention heads.

Previous work has attempted to address this issue by proposing to share projection matrices or eliminating them entirely to improve the parameter efficiency of \textsc{MHA}. \citet{Lan2020ALBERT:} proposed sharing projection parameters for keys, queries and values across layers, while \citet{Kitaev2020Reformer:} introduced a method for sharing the projection matrix between keys and values within each transformer layer. Additionally, similar approaches use a multi-query attention approach that uses a pair of global projection matrices for keys and values in each layer~\citep{shazeer2019fast, chowdhery2022palm, ainslie2023colt5}. Furthermore, \citet{yan2021attention} eliminate the projection matrices entirely and directly treat the input hidden states as both keys and values. In a different direction, ~\citet{lee-thorp-etal-2022-fnet} propose models that replace the attention blocks with token-mixture blocks (i.e. using linear or Fourier transformations) that contain fewer or no parameters compared to \textsc{MHA}.

Inspired by the position embeddings in transformers~\citep{vaswani2017attention,devlin-etal-2019-bert}, we aim to simplify and reduce the memory footprint of the \textsc{MHA} mechanism. We achieve this using only a single projection matrix for each of the keys, queries and values respectively shared across all attention heads, and one embedding per head (\textsc{MHE}). 

Our contributions are as follows:

\begin{itemize}
    \item We propose \textsc{MHE}, a novel attention module that uses shared projection matrices across heads that are modified by corresponding embedding heads. Our method generates multiple attention heads requiring only a small fraction of additional parameters compared to single-head attention.

    \item We empirically demonstrate that our \textsc{MHE} attention is substantially more parameter efficient compared to alternative attention mechanisms while achieving high predictive performance retention ratio (i.e. 92.9\textasciitilde 98.7\%) to \textsc{MHA} on several downstream tasks. \textsc{MHE} is $(3n^2-3n)d^2-3nd$ smaller than \textsc{MHA} for a single attention sublayer with $n$ attention heads and a hidden dimension of $d$ per head.
\end{itemize}

\section{Related Work}

\subsection{Model Compression}

To make PLMs memory efficient, previous work has focused on the following post-hoc model compression approaches~\citep{ganesh-etal-2021-compressing,tay2022efficient}.

\paragraph{Quantization} \citet{hubara2017quantized} proposed representing weights using fewer bits to reduce memory requirements. \citet{zadeh2020gobo} introduced a method for identifying the outliers in weights and excluded them during quantization. Another direction involves additional training steps to adjust the quantized weights, i.e. quantization-aware training \citep{zafrir2019q8bert, boo2020fixed, stock2020training, shen2020q, tambe2021edgebert, tao-etal-2022-compression}. \citet{bai2022towards} developed a more efficient post-training quantization approach that minimizes the reconstruction error incurred by quantization.

\paragraph{Pruning} These compression approaches remove entirely parts of the network such as weights close to zero~\citep{gordon-etal-2020-compressing, mao-etal-2020-ladabert, chen2020lottery} and weights that move towards zero during fine-tuning~\citep{sanh2020movement, tambe2021edgebert}. Different to operating on individual weights, previous work attempted to remove structured blocks of weights or even architectural components such as attention heads and encoder layers~\citep{fan2019reducing, prasanna-etal-2020-bert, khetan-karnin-2020-schubert, li-etal-2020-efficient-transformer, lin-etal-2020-pruning, tay2021synthesizer}.

\paragraph{Knowledge Distillation} This set of techniques typically train a light-weight student model to mimic the outputs of a larger teacher PLM~\citep{sun-etal-2019-patient,li-etal-2020-bert,jiao-etal-2020-tinybert,sun-etal-2020-mobilebert,li-etal-2021-dynamic,tahaei-etal-2022-kroneckerbert}. In a similar direction, smaller PLMs have been recently fine-tuned on text generated by larger PLMs~\citep{chiang2023vicuna,alpaca}.

\paragraph{Weight Matrix Decomposition} Previous work also proposed replacing large weight matrices by the product of two smaller ones for reducing model size and runtime memory. Weight matrix decomposition has been applied to linear layers~\citep{mao-etal-2020-ladabert,ben-noach-goldberg-2020-compressing}, the embedding matrix~\citep{lan2020albert, tambe2021edgebert, wang2022exploring}, and attention blocks~\citep{hu2022lora,wang2022exploring}.

\paragraph{Embedding Matrix Compression}
Finally, various attempts have been introduced for compressing the embedding matrix during pre-training and fine-tuning~\citep{xue-etal-2022-byt5,clark-etal-2022-canine,xue-aletras-2022-hashformers}.

\subsection{Improving Attention Efficiency}

Previous work on making attention more efficient includes efforts towards (1) speeding-up pairwise computations between token representations; and (2) parameter efficiency.

\paragraph{Computational Efficiency} 
While improving computational efficiency of attention is out of the scope of our paper, we provide a brief overview of previous work since it is complementary to parameter efficiency. 
One approach to speed up attention computation is by reducing the number of similarity computations between representations in different positions using predefined local windows, fixed or dynamic strides~\citep{child2019generating, zaheer2020big, beltagy2020longformer, Kitaev2020Reformer:}. Other methods leverage the approximation of SoftMax to change the order of matrix multiplications, resulting in lower computational complexity~\citep{katharopoulos2020transformers,choromanski2021rethinking,schlag2021linear,zhen2022cosformer}. Related approaches along this direction proposed kernel functions that require additional parameters \citep{choromanski2021rethinking,wang2020linformer}.
Finally, \citet{dao2022flashattention} proposed improvements in GPU memory access to optimize and accelerate the \textsc{MHA} computation.

\paragraph{Memory Efficiency} \citet{Lan2020ALBERT:} introduced a method for sharing the projection parameters for queries, keys and values across transformer layers. Furthermore, \citet{Kitaev2020Reformer:} proposed sharing the projection matrix between keys and values within each layer. Additionally, other methods use a multi-query attention approach that shares projection weights for keys and values across different heads~\citep{shazeer2019fast, chowdhery2022palm, ainslie2023colt5}, while \citet{yan2021attention} directly treat the input hidden states as both keys and values. In a different direction, ~\citet{lee-thorp-etal-2022-fnet} proposed replacing the attention blocks with faster token-mixture blocks consisting of a few parameters or no parameters at all.
This includes methods such as linear or Fourier transformations in the token-mixture block. However, these  approaches tend to yield lower predictive performance compared to \textsc{MHA}.

\section{Multiple Head Embeddings Attention}

Inspired by the absolute position embeddings~\citep{vaswani2017attention,devlin-etal-2019-bert} for distinguishing the representation of the same token in different contexts, we propose Multiple Head Embeddings (\textsc{MHE}) attention. \textsc{MHE} uses a shared `seed' projection matrix that is subsequently combined with distinct head embeddings to generate multiple attention heads. 

\subsection{Multi-head Attention (MHA)}
We first begin by formally defining \textsc{MHA}. \textsc{MHA} consists of different projection matrices ($\mathbf{W}^Q_i, \mathbf{W}^K_i, \mathbf{W}^V_i\in\mathbb{R}^{d_m\times d_h}, i=1,...,n$, where $d_m$ is the dimension of the input representation and $d_h$ is the dimension of $n$ attention heads) for queries ($Q$), keys ($K$) and values ($V$) per head, $3\times n$ in total. It is computed as follows:

{\small
  \begin{align}
    \mathbf{Q}_i,\mathbf{K}_i,\mathbf{V}_i&=\mathbf{X}\mathbf{W}_i^{Q,K,V}\label{eq:proj}\\
   \mathbf{H}_i&=\text{Att}(\mathbf{Q}_i,\mathbf{K}_i,\mathbf{V}_i)\\&=\text{SoftMax}(\frac{\mathbf{Q}_i\mathbf{K}_i^\top}{\sqrt{d_h}})\mathbf{V}_i\label{eq:att}
  \end{align}
}
Note that we use scale-dot attention, but our method can be used with any other attention mechanism.

\subsection{Seed Projection Matrix}
Unlike \textsc{MHA} that uses different projection matrices per head, \textsc{MHE} attention employs only a single projection matrix for each of the queries, keys and values, $\mathbf{W}^Q, \mathbf{W}^K, \mathbf{W}^V\in\mathbb{R}^{d_m\times d_h}$. These matrices are shared across all attention heads. 

We obtain query, key and values projections of the input sequence $\textbf{X}$ as follows:

{\small
\begin{align}
    \mathbf{Q,K,V} &=\mathbf{X}\mathbf{W}^{Q,K,V}\label{eq:mhe_proj}
\end{align}
}

\subsection{Attention Head Embeddings}
Using a seed projection matrix for $\mathbf{Q,K,V}$ is equivalent to a single-head attention (\textsc{SHA}) module. Therefore, we need a mechanism to transform the seed projection matrices to obtain different attention head. For this purpose, we represent each attention head $i$ by specific head embeddings $\mathbf{e}_i^Q, \mathbf{e}_i^K, \mathbf{e}_i^V\in\mathbb{R}^{d_h}, i=1,...,n$ for queries, key and values. These embeddings have a substantially smaller memory footprint compared to using different projection matrices per head. The contextualized representation $\mathbf{H}_i$ of the entire input sequence $\mathbf{X}$ for head $i$ is computed as follows:

{\small
    \begin{align}\mathbf{\tilde{Q}}_i,\mathbf{\tilde{K}}_i,\mathbf{\tilde{V}}_i&=\psi(\mathbf{Q;K;V},\mathbf{e}_i^{Q,K,V})\label{eq:mhe_head}\\
   \mathbf{H}_i&=\text{Att}(\mathbf{\tilde{Q}}_i,\mathbf{\tilde{K}}_i,\mathbf{\tilde{V}}_i)\label{eq:mhe_att}
  \end{align}
}
\noindent where $\psi(\cdot )$ is a function that modifies the query, key and value matrices with a corresponding head embedding $\mathbf{e}_i$.

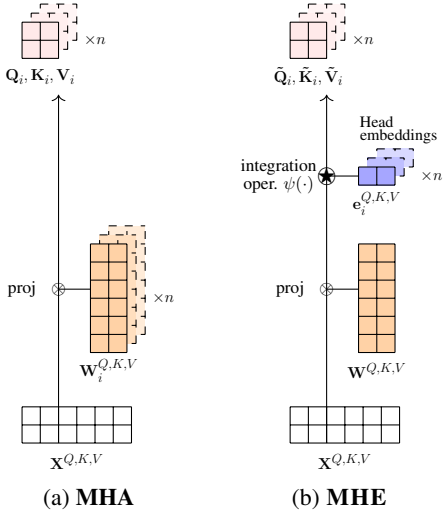
\begin{figure}[!t]
    \centering

    \begin{subfigure}[b]{0.2\textwidth}
        \centering
        \begin{tikzpicture}[scale=0.6,transform shape]
            \begin{scope}[shift={(0mm,0mm)}, scale=0.4]
                \node at (3,-1) {\Huge$\mathbf{X}^{Q,K,V}$};
                \draw (0,0) grid (6,2);
            \end{scope}
            
            \begin{scope}[shift={(19mm,24mm)}, scale=0.4]
                \fill[orange!15] (0,0) rectangle ++(2,6);
                \draw[dashed] (0,0) grid (2,6);
            \end{scope}
            \begin{scope}[shift={(17mm,22mm)}, scale=0.4]
                \fill[orange!25] (0,0) rectangle ++(2,6);
                \draw[dashed] (0,0) grid (2,6);
            \end{scope}
            \begin{scope}[shift={(15mm,20mm)}, scale=0.4]
                \node at (4,3) {\Huge$\times n$};
                \node at (1,-1) {\Huge$\mathbf{W}_i^{Q,K,V}$};
                \fill[orange!35] (0,0) rectangle ++(2,6);
                \draw (0,0) grid (2,6);
            \end{scope}

            \begin{scope}[shift={(4mm,89mm)}, scale=0.4]
                \fill[pink!15] (0,0) rectangle ++(2,2);
                \draw[dashed] (0,0) grid (2,2);
            \end{scope}
            \begin{scope}[shift={(2mm,87mm)}, scale=0.4]
                \fill[pink!25] (0,0) rectangle ++(2,2);
                \draw[dashed] (0,0) grid (2,2);
            \end{scope}
            \begin{scope}[shift={(0mm,85mm)}, scale=0.4]
                \node at (4,1) {\Huge$\times n$};
                \node at (1,-1) {\Huge$\mathbf{Q}_i,\mathbf{K}_i,\mathbf{V}_i$};
                \fill[pink!35] (0,0) rectangle ++(2,2);
                \draw (0,0) grid (2,2);
            \end{scope}

            \draw[->] (0.8, 0) -> (0.8,7.7);
            \draw[-] (1.5, 3.4) -> (0.8,3.4);
            \node at (0.8, 3.4) {\large$\otimes$};
            \node at (0, 3.4) {proj};
    
        \end{tikzpicture}
        \caption{\textbf{MHA}}
        \label{fig:proj_mha}
    \end{subfigure}
    \begin{subfigure}[b]{0.2\textwidth}
        \centering
        \begin{tikzpicture}[scale=0.6,transform shape]

            \begin{scope}[shift={(0mm,0mm)}, scale=0.4]
                \node at (3,-1) {\Huge$\mathbf{X}^{Q,K,V}$};
                \draw (0,0) grid (6,2);
            \end{scope}

            \begin{scope}[shift={(15mm,20mm)}, scale=0.4]
                \node at (1,-1) {\Huge$\mathbf{W}^{Q,K,V}$};
                \fill[orange!35] (0,0) rectangle ++(2,6);
                \draw (0,0) grid (2,6);
            \end{scope}

            \begin{scope}[shift={(19mm,61mm)}, scale=0.4]
                \fill[blue!15] (0,0) rectangle ++(2,1);
                \draw[dashed] (0,0) grid (2,1);
            \end{scope}
            \begin{scope}[shift={(17mm,59mm)}, scale=0.4]
                \fill[blue!25] (0,0) rectangle ++(2,1);
                \draw[dashed] (0,0) grid (2,1);
            \end{scope}
            \begin{scope}[shift={(15mm,57mm)}, scale=0.4]
                \node at (4,0.5) {\Huge$\times n$};
                \node at (1,-1) {\Huge$\mathbf{e}_i^{Q,K,V}$};
                \node at (1,3.5) {\Huge Head};
                \node at (2.2,2.7) {\Huge embeddings};
                \fill[blue!35] (0,0) rectangle ++(2,1);
                \draw (0,0) grid (2,1);
            \end{scope}

            \begin{scope}[shift={(4mm,89mm)}, scale=0.4]
                \fill[pink!15] (0,0) rectangle ++(2,2);
                \draw[dashed] (0,0) grid (2,2);
            \end{scope}
            \begin{scope}[shift={(2mm,87mm)}, scale=0.4]
                \fill[pink!25] (0,0) rectangle ++(2,2);
                \draw[dashed] (0,0) grid (2,2);
            \end{scope}
            \begin{scope}[shift={(0mm,85mm)}, scale=0.4]
                \node at (4,1) {\Huge$\times n$};
                \node at (1,-1) {\Huge$\mathbf{\tilde{Q}}_i,\mathbf{\tilde{K}}_i,\mathbf{\tilde{V}}_i$};
                \fill[pink!35] (0,0) rectangle ++(2,2);
                \draw (0,0) grid (2,2);
            \end{scope}

            \draw[->] (0.8, 0) -> (0.8,7.7);
            \draw[-] (1.5, 3.4) -> (0.8,3.4);
            \node at (0.8, 3.4) {\large$\otimes$};
            \node at (0, 3.4) {proj};

            \node at (-0.25, 6.1) {integration};
            \node at (-0.25, 5.7) {oper. $\psi(\cdot)$};
            \node at (0.8, 5.9) {\large\textcircled{$\bigstar$}};
            \draw[-] (1.5, 5.9) -> (0.8,5.9);
    
        \end{tikzpicture}
        \caption{\textbf{\textsc{MHE}}}
        \label{fig:proj_ohhe}
    \end{subfigure}
    \caption{Multi-head attention (left) requires $3\times n$ projection matrices for queries, keys and values ($\mathbf{W}^{Q,K,V}$) where $n$ is the number of attention heads. Multi-head embedding attention (right) uses only three projection matrices and $3\times n$ head embeddings. }
    \label{fig:proj_compare}
\end{figure}

\subsection{Modifying Queries, Keys and Values with Head Embeddings}
We propose two \textsc{MHE} variants, one adds and the other multiplies the head embeddings with the seed projection matrices.

\paragraph{\textsc{MHE-Add}:} 
Motivated by the absolute position embedding~\citep{devlin-etal-2019-bert}, we use the addition operation in Equation \ref{eq:mhe_head}, represented as $\psi(\mathbf{A}, \mathbf{b}):=\mathbf{A}+\mathbf{b}$, where $\mathbf{A} \in \{\mathbf{Q},\mathbf{K},\mathbf{V}\}$ and $\mathbf{b} \in \{\mathbf{e}^Q,\mathbf{e}^K,\mathbf{e}^V\}$ respectively.

\paragraph{\textsc{MHE-Mul}:} Likewise, motivated by the rotary position embedding~\citep{su2021roformer}, \textsc{MHE-Mul} employs multiplication as the integrating operation in Equation \ref{eq:mhe_head} as $\psi(\mathbf{A}, \mathbf{b}):=\mathbf{A}\odot(\mathbf{b}+1)$, where $\odot$ represents the Hadamard product.\footnote{We add 1 to avoid elements in queries, keys and values become too small during initialization.}

Figure \ref{fig:proj_compare} shows an overview of the \textsc{MHE} mechanism compared to \textsc{MHA}.

\section{Experimental Setup}

\subsection{Attention Mechanisms}
We compare our \textsc{MHE} attention with the following attention mechanisms:\footnote{We have also experimented with Linear and Fourier token-mixture models~\citep{lee-thorp-etal-2022-fnet} yielding substantially lower performance. For full results of these methods, see~\autoref{sec:metrics}.}

\begin{itemize}

\item\textbf{Multi-head Attention (\textsc{MHA}):} This is the original multi-head attention mechanism~\citep{vaswani2017attention,devlin-etal-2019-bert}.

\item \textbf{Single-head Attention (\textsc{SHA}):} Similar to \textsc{MHA} but using only one attention head.

\item \textbf{\textsc{EL-att}:} Introduced by \citet{yan2021attention}, this attention variant completely eliminates the projection matrices for all keys and values.

\item \textbf{\textsc{MQA}:} Introduced by \citet{shazeer2019fast}, this approach uses shared projection matrices for keys and values across all attention heads. Note that different projection matrices are used for queries across heads.

\item \textbf{\textsc{SKV}:} Introduced by \citet{Kitaev2020Reformer:}, this attention variant enforces keys and values to share the same projection matrix within each attention module.
\end{itemize}

\subsection{Data} 
We experiment with a diverse range of tasks including: (1) two standard natural language understanding benchmarks in English, \textsc{Glue} \citep{wang-etal-2018-glue} and \textsc{SuperGlue} \citep{wang2019superglue}; (2) two question and answering benchmarks in English, \textsc{SQuAD v1.1} \citep{rajpurkar-etal-2016-squad} and \textsc{SQuAD v2.0} \citep{rajpurkar-etal-2018-know}; (3) \textsc{WMT-14} English-to-German machine translation \citep{bojar-etal-2014-findings}; and (4) two language modelling datasets in English \textsc{WikiText-103} \citep{merity2017pointer} and \textsc{Penn Treebank} \citep{marcus-etal-1993-building}.

\subsection{Models}
We test all different attention variants on two architectures: (1) encoder-only transformer~\citep{devlin-etal-2019-bert} and (2) encoder-decoder transformer~\citep{vaswani2017attention}. 

\paragraph{Encoder-only}
For \textsc{Glue}, \textsc{SuperGlue}, \textsc{SQuAD v1.1} and \textsc{SQuAD v2.0}, we use a \textsc{BERT}-base architecture. This consists of 12 transformer layers, embedding size of 768, hidden states dimension of 768, 12 attention heads and a maximum sequence length of 512.

\paragraph{Decoder-only}
We also test a decoder-only model using the GPT2-base architecture on \textsc{WikiText-103}, \textsc{Penn Treebank} and \textsc{Glue}. GPT2-base consists of 12 transformer layers, embedding size of 768, hidden states dimension of 768, 12 attention heads and a maximum sequence length of 512.

\paragraph{Encoder-decoder}
For \textsc{WMT-14}, we train an encoder-decoder transformer from scratch. It consists of 12 layers (6 for the encoder and decoder respectively), an embedding size of 512, hidden states dimension of 512 and 8 attention-heads and a maximum sequence length of 100.

We set the number of attention heads to 1 for all \textsc{SHA} models. Experimenting with larger models and different number of attention heads is out of the scope of our paper and left for future work due to limited access to computing resources.

\subsection{Implementation Details}

\paragraph{Pre-training}
We pre-train all models on the English Wikipedia and BookCorpus ~\citep{zhu2015aligning} from HuggingFace ~\citep{lhoest-etal-2021-datasets} for up to 1M steps with a batch size of 128. We choose masked language modelling as the pre-training objective. For all models, we use a 30K WordPiece vocabulary~\citep{devlin-etal-2019-bert}.

\paragraph{Fine-tuning and Training}
For \textsc{Glue}, \textsc{SuperGLue}, \textsc{SQuAD v1.1} and \textsc{SQuAD v2.0}, we fine-tune all pre-trained models up to 20 epochs with early stopping fixing the batch size to 32. For each task, we use five different seeds and report the average. 

We train the encoder-decoder model from scratch on the training set of \textsc{WMT-14} English-to-German machine translation dataset up to 100K steps with a batch size of 256. \textsc{WMT-14} contains 4.5M sentence pairs and evaluate on its test set. We train the tokenizer using byte-pair-encoding \citep{sennrich-etal-2016-neural} with 37K merging steps on the training set. We enable both source language and target language to share the vocabulary.
We use one random seed and report the average on the last five epochs.
We optimize all models using AdamW~\citep{loshchilov2018decoupled}.

\begin{table*}[!t]
\small
\begin{center}
\resizebox{\linewidth}{!}{
\strutlongstacks{T}
\begin{tabular}{l|r|crr:crr:crr:crr}
\toprule
 &  & \multicolumn{3}{c:}{\textbf{\textsc{Glue}}} & \multicolumn{3}{c:}{\textbf{\textsc{SuperGlue}}} & \multicolumn{3}{c:}{\textbf{\textsc{SQuAD} v1.1}} & \multicolumn{3}{c}{\textbf{\textsc{SQuAD v2.0}}}\\
\textbf{Attention} & \makecell[c]{\textbf{\#params}} & Acc & PRR & PEoP & Acc & PRR & PEoP & Acc & PRR & PEoP & Acc & PRR & PEoP\\\midrule
\textsc{SHA} & 8.85M & 79.2&96.7& - & 67.1&95.1& - & 82.5&93.1& - & 67.6&87.8& -\\ 
\textsc{MHA} & 28.32M & 81.9&100.0&0.02 & 70.5&100.0&0.02 & 88.6&100.0&0.03 & 77.0&100.0&0.06\\\midrule
\textsc{EL-att} & 14.16M & 80.3&98.0&0.02 & 69.5&98.5&0.06 & 86.5&97.6&0.08 & 72.2&93.8&0.11\\
\textsc{MQA} & 15.34M & 81.3&99.2&0.04 & 69.3&98.2&0.04 & 86.7&97.9&0.07 & 74.8&97.1&0.15\\
\textsc{SKV} & 21.23M & \textbf{81.4}&\textbf{99.4}&0.02 & \textbf{69.9}&\textbf{99.1}&0.03 & \textbf{88.1}&\textbf{99.4}&0.05 & \textbf{75.9}&\textbf{98.6}&0.09\\
\hdashline
\textsc{MHE-Add} & 8.88M & 80.4&98.2&4.92 & 69.1&97.9&9.44 & 83.7&94.5&4.65 & 71.8&93.2&19.88\\
\textsc{MHE-Mul} & 8.88M & 80.6&98.3&\textbf{5.53} & 69.6&98.7&\textbf{12.07} & 85.9&97.0&\textbf{13.19} & 72.3&93.9&\textbf{22.25}\\

\bottomrule
\end{tabular}
}
\caption{Results of the encoder-only architecture on \textsc{Glue}, \textsc{SuperGlue}, \textsc{SQuAD v1.1} and \textsc{SQuAD v2.0} dev sets with performance retention ratio (PRR) and performance elasticity of parameters (PEoP)  over five runs. \textbf{Bold} values denote best performing method in each benchmark. }

\label{table:perform_result}
\end{center}
\end{table*}

\begin{table*}[!t]
\begin{center}
\small
\setlength{\tabcolsep}{5pt}

\strutlongstacks{T}
\begin{tabular*}{\textwidth}{@{\extracolsep{\fill}\quad}l|r|crr:crr:crr}
\toprule
 &  & \multicolumn{3}{c:}{\textbf{\textsc{Glue}}} & \multicolumn{3}{c:}{\textbf{\textsc{WikiText-103}}} & \multicolumn{3}{c}{\textbf{\textsc{Penn Treebank}}} \\
\textbf{Attention} & \makecell[c]{\textbf{\#params}} & Acc & PRR & PEoP & PPL & PRR & PEoP & PPL & PRR & PEoP \\\midrule
\textsc{SHA} & 8.85M & 75.3 & 97.2 & - & 62.0 & 55.8 & - & 68.1 & 46.3 & -\\ 
\textsc{MHA} & 28.32M & 77.5 & 100.0 & 0.01 & 43.0 & 100.0 & 0.14 & 44.3 & 100.0	& 0.16 \\\midrule
\textsc{EL-att} & 14.16M & 76.6 & 98.9 & 0.03 & 57.1 & 67.2 & 0.13 & 56.1 & 73.4 & 0.29 \\
\textsc{MQA} & 15.34M & 76.9 & 99.2 & 0.03 & 49.7 & 84.4 & 0.27 & 49.3 & 88.7 &0.38 \\
\textsc{SKV} & 21.23M & \textbf{77.1} & \textbf{99.5} & 0.02 & \textbf{46.2} & \textbf{92.6} & 0.18 & \textbf{45.5} & \textbf{97.3} & 0.24 \\
\hdashline
\textsc{MHE-Add} & 8.88M & 75.8 & 97.8 & 2.18 & 54.0 & 74.4 & 41.29 & 55.3 & 75.2 & 60.15 \\
\textsc{MHE-Mul} & 8.88M & 76.7 & 99.0 & \textbf{5.92} & 53.8 & 74.9 & \textbf{42.32} & 50.7 & 85.6 & \textbf{81.76} \\

\bottomrule
\end{tabular*}
\caption{Results of decoder-only architecture on \textsc{Glue} dev sets and \textsc{WikiText-103}, \textsc{Penn Treebank} test sets with performance retention ratio (PRR) and performance elasticity of parameters (PEoP)  over five runs. \textbf{Bold} values denote best performing method in each benchmark.}

\label{table:perform_result_gpt}
\end{center}
\end{table*}

\paragraph{Hyperparameters}
Hyperparameter selection details are in \autoref{sec:hyperparameters}.

\paragraph{Hardware}
For pre-training, we use four NVIDIA Tesla A100 GPUs and one for fine-tuning on downstream tasks.

\subsection{Predictive Performance Evaluation} \label{sec:Evaluation}
For \textsc{Glue}, \textsc{SuperGlue}, \textsc{SQuAD v1.1} and \textsc{SQuAD v2.0}, we use the official metric of each task (see \autoref{sec:metrics} for details on metrics for each task). We report F1 score for \textsc{SQuAD v1.1} and \textsc{SQuAD v2.0}. We use BLEU to report performance in \textsc{WMT-14} English-to-German machine translation task. We use perplexity (PPL) to report generative performance on \textsc{WikiText-103} and \textsc{Penn Treebank} by fixing the stride length to 256.

\subsection{Memory Efficiency Evaluation}\label{sec:Efficiecy}
Furthermore, we use the following metrics to measure and compare the memory efficiency of \textsc{MHE} and the baselines.

\begin{itemize}
\item\textbf{Performance Retention Ratio:} We compute the ratio between the predictive performance of each attention mechanism compared to \textsc{MHA}  upper-bound baseline performance (the higher the better).

For direct indicator (e.g. accuracy etc.):
{\small
\begin{equation*}
\text{PRR} = \frac{\text{score}_ \text{model}}{\text{score}_{\text{MHA}}}
\end{equation*}
}

For inverse indicator (e.g. perplexity etc.):
{\small
\begin{equation*}
\text{PRR} = 1-\frac{\text{score}_ \text{model}-\text{score}_{\text{MHA}}}{\text{score}_{\text{MHA}}}
\end{equation*}
}

\item\textbf{Performance Elasticity of Parameters:}
Inspired by the concept of elasticity in economics~\citep{bittermann1934elasticity}, which measures the responsiveness of an economic variable (e.g. investment demand) to a change in another (e.g. interest rate), we extend it to measure the parameter utilization rate of a target model compared to the \textsc{SHA} lower-bound. The performance elasticity of parameters (PEoP) indicates how effectively parameters contribute to predictive performance, compared to \textsc{SHA}. It is computed as follows:

For direct indicator (e.g. accuracy etc.):
{\small
\begin{equation*}
\text{PEoP}=\frac{(\text{score}_\text{model}/\text{score}_{\text{SHA}})-1}{(\text{params}_\text{model}/\text{params}_{\text{SHA}})-1}
\end{equation*}
}

For inverse indicator (e.g. perplexity etc.):
{\small
\begin{equation*}
\text{PEoP}=-\frac{(\text{score}_\text{model}/\text{score}_{\text{SHA}})-1}{(\text{params}_\text{model}/\text{params}_{\text{SHA}})-1}
\end{equation*}
}

PEoP quantifies the extent to which a model's performance can be boosted with 1\% additional parameters compared to a baseline model (the higher the better).\footnote{We subtract 1 in both nominator and denominator, following the original definition of elasticity.}

\end{itemize}

\section{Results} \label{sec:Results}

\subsection{Predictive Performance Comparison}

Table \ref{table:perform_result} presents results on \textsc{Glue}, \textsc{SuperGlue}, \textsc{SQuAD v1.1} and \textsc{SQuAD v2.0} for our \textsc{MHE} variants and all baselines. We first observe that both the performance of our \textsc{MHE-Add} and \textsc{MHE-Mul} are comparable to the vanilla \textsc{MHA} on two text classification benchmarks (80.4, 80.6 vs. 81.9 on average \textsc{Glue} and 69.1, 69.6 vs. 70.5 on average \textsc{SuperGlue}) with high performance retention ratios (PRR) between 97.9\% and 98.7\%. On question answering tasks \textsc{SQuAD v1.1} and \textsc{SQuAD v2.0}, both \textsc{MHE} variants are also competitive, with PRRs higher than 93\%.

Similar results are observed on the \textsc{WMT-14} English-to-German machine translation task for the encoder-decoder transformer. According to Table \ref{table:seq2seq_result}, \textsc{MHE-Add} and \textsc{MHE-Mul} achieve BLEU scores of 23.0 and 23.6, respectively. The performance of \textsc{MHE-Mul} is negligibly lower than that of \textsc{MHA} (24.8) while being substantially smaller.

Consistent results for the decoder-only transformer are shown in Table \ref{table:perform_result_gpt}. The PRRs for \textsc{MHE-Add} and \textsc{MHE-Mul} on \textsc{Glue} are still high (i.e. 97.8\% and 99.0\%). While using the intrinsic metrics for evaluation, \textsc{MHE-Mul} leads to the perplexities of 53.8 and 50.7 compared to 43.0 and 44.3 for \textsc{MHA} on \textsc{WikiText-103} and \textsc{Penn Treebank} respectively, indicating a stable PRR higher than 74.9\%.

In all tasks, \textsc{MHE} consistently outperforms SHA by a large margin with only 0.03M extra parameters, i.e. 0.6\textasciitilde 17.4. For example,  69.6 vs. 67.1 in \textsc{SuperGlue}, 72.3 vs. 67.6 in \textsc{SQuAD v2.0}, 23.6 vs. 22.5 in \textsc{WMT-14} and 62.0 vs. 53.8 in \textsc{WikiText-103} for the \textsc{MHE-Mul} variant. We also note that \textsc{MQA} and \textsc{SKV} attention mechanisms generally perform better than \textsc{MHE}, however they are 1.7 and 2.4 times larger than \textsc{MHE}, i.e. 15.34M and 21.23M vs. 8.88M parameters. It is worth noting that \textsc{MHE-Mul} outperforms \textsc{El-Att} on three out of five benchmarks, despite having nearly half the parameters in the attention module.

\subsection{Memory Efficiency Comparison}

Our results so far indicate that performance increases with the number of attention mechanism parameters, which is expected. Next, we inspect how efficiently different attention mechanisms utilize their parameters \footnote{For a detailed report on the memory usage of different attention mechanisms, see Appendix \ref{sec:memory_usage}.}. Tables \ref{table:perform_result} and \ref{table:seq2seq_result} show how parameter efficient our two \textsc{MHE} attention variants and all baselines are, measured in PEoP. Note that PEoP scores for \textsc{SHA} cannot be computed as it is used as the point for reference model. We also report PRR using \textsc{MHA} as a baseline for completeness, however this metric does not take the model size into account.

We first observe in Table \ref{table:perform_result} that both our \textsc{MHE-Add} and \textsc{MHE-Mul} achieve the highest PEoP scores on the two natural language understanding benchmarks (4.92, 5.53 on \textsc{Glue}, and 9.44, 12.07 on \textsc{SuperGlue}) and two question answering tasks (4.65, 13.19on \textsc{SQuAD v1.1}, and 19.88, 22.25 on \textsc{SQuAD v2.0}). In contrast, vanilla \textsc{MHA} results in the lowest PEoP score among all models as expected, ranging from 0.02 to 0.06. It indicates the memory inefficiency of \textsc{MHA}.

The PEoPs of more light-weight \textsc{El-Att} and \textsc{SKV} are similar to that of \textsc{MHA} (0.02) on average \textsc{Glue}, barely 4~\textperthousand  of that of \textsc{MHE}, indicating they are far more memory-inefficient compared to \textsc{MHE}.

Similar findings are observed in \textsc{WMT-14} for the encoder-decoder models depicted in Table \ref{table:seq2seq_result}. \textsc{MHE-Add} and \textsc{MHE-Mul} achieve PEoP scores of 20.0 and 27.9, respectively. In contrast, the PEoP scores of \textsc{MHA}, \textsc{El-Att} \textsc{MQA} and \textsc{SKV} are close to zero (barely 0.1). This means that investing more parameters into their attention modules would not bring proportional benefits in predictive performance. Even for the \textsc{SKV} which is half the size of \textsc{MHA} and achieves high PRR, when the number of parameters increase by 1\%, the BLEU score increases a negligible 0.1\%, while evolving from \textsc{SHA}. However, with the same number of parameters, our most memory-inefficient \textsc{MHE-Mul} is able to improve the BLEU score by 11.0\%. Such rate of return is 110 times larger than that of \textsc{SKV}. Leveraging the head embeddings by adding only a negligible number of parameters efficiently improves the predictive performance.

We further observe that \textsc{MHE-Add} and \textsc{MHE-Mul} are architecture-agnostic, obtaining similar memory efficiency for the decoder-only model in Table \ref{table:perform_result_gpt}. Both our \textsc{MHE-Add} and \textsc{MHE-Mul} achieve the highest PEoP scores on the two language modelling benchmarks (41.29, 42.32 on \textsc{WikiText-103} and 60.15 and 81.76 on \textsc{Penn Treebank}) and \textsc{Glue} (2.18 and 5.92). At the same time, \textsc{MHA} fail to perform well on \textsc{Glue} and \textsc{Penn Treebank} with a PEoP of 0.01 and 0.16 respectively. \textsc{MHE-Add} and \textsc{MHE-Mul} also consistently outperform other efficient-attention variants (i.e. \textsc{EL-Att}, \textsc{MQA} and \textsc{SKV}) by 72\textasciitilde 340 times on PEoP across the three benchmarks.

In all tasks, \textsc{MHE} consistently outperforms \textsc{MHA} by orders of magnitude in parameter efficiency. We also note that \textsc{El-Att}, \textsc{MQA} and \textsc{SKV} only lead to PEoP scores with the same magnitude as \textsc{MHA}. This highlights the more superior parameter utilization of \textsc{MHE} attention variants, achieving state-of-the-art memory-efficiency.

\begin{table}[!t]
\begin{center}
\small
\setlength{\tabcolsep}{5pt}
\strutlongstacks{T}
\begin{tabular*}{\linewidth}{@{\extracolsep{\fill}\quad}l|r|crr}
\toprule

\textbf{Attention} & \makecell[c]{\textbf{\#params}} & \makecell[c]{\textbf{BLEU}} & \makecell[c]{\textbf{PRR}} & \makecell[c]{\textbf{PEoP}}\\\midrule
\textsc{SHA} & 6.49M & 22.5 & 90.8 & -\\
\textsc{MHA} & 18.87M & 24.8 & 100.0 & 0.1\\ \midrule
\textsc{EL-att} & 9.44M & 23.9 & 96.6 & 0.1\\
\textsc{MQA} & 10.62M & 24.2 & 97.6 & 0.1\\
\textsc{SKV} & 14.16M & \textbf{24.7} & \textbf{99.5} & 0.1\\\hdashline
\textsc{MHE-Add} & 6.52M & 23.0 & 92.9 & 5.5\\
\textsc{MHE-Mul} & 6.52M & 23.6 & 95.0 & \textbf{11.0}\\

\bottomrule
\end{tabular*}

\caption{BLEU scores on \textsc{WMT-14} English to German machine translation task with performance retention ratio (PRR) and performance elasticity of parameters (PEoP). \textbf{Bold} values denote best performing method in each benchmark.} 

\label{table:seq2seq_result}
\end{center}
\end{table}

\subsection{Theoretical Memory Complexity}
Table \ref{table:save_result} presents the theoretical memory complexity and the total number of parameters of our two \textsc{MHE} and baseline attention mechanisms in a single transformer sublayer. First, we see that the theoretical memory complexity of \textsc{MHA} and other efficient parameters (\textsc{El-Att}, \textsc{MQA} and \textsc{SKV}) are quadratic with the number of attention heads, while our \textsc{MHE} are the only two variants having the complexity linear with the attention heads similar to \textsc{SHA}.

Taking a closer look at the rightmost column in Table \ref{table:save_result}, we observe that the number of extra parameters of all attention variants compared to \textsc{SHA} have a quadratic relationship to both the number $n$ and the dimension of attention heads $d$, except our two \textsc{MHE} variants. \textsc{MHE} only requires a relatively small fraction of additional parameters compared to \textsc{SHA}.

\subsection{Scaling the Number of Attention Parameters}

\begin{figure}[!t]
    \centering
    \includegraphics[width=0.97\linewidth]{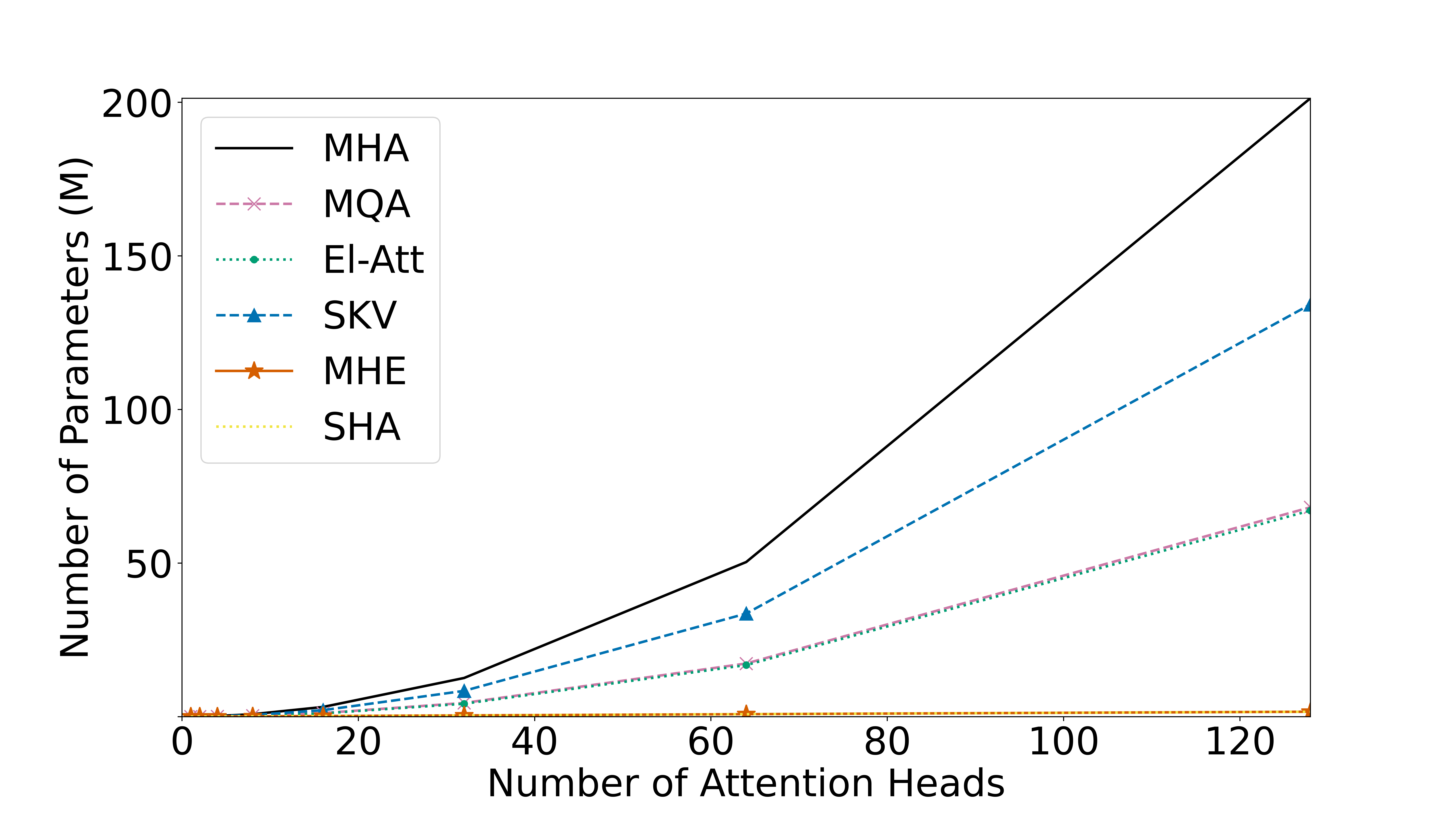}
    \caption{Number of parameters per attention sublayer, while scaling the number of attention heads in different attention variants. We fix the dimension of attention to 64.
    }
    \label{fig:scaling_params}
\end{figure}

Delving deeper to the effect of scaling to memory footprint, we show in Figure \ref{fig:scaling_params} the total number of parameters needed for a single attention module (e.g. in an encoder layer). We fix the dimension of attention heads to 64 commonly used by BERT~\citep{devlin-etal-2019-bert}, RoBERTa~\citep{liu2019roberta}, GPT-2~\citep{radford2019language}, BART~\citep{lewis-etal-2020-bart} and T5~\citep{raffel2020exploring}.
In general, we note that the number of parameters in \textsc{MHA} could reach more than 200M if employing 128 attention heads. At the same time, \textsc{SKV}, \textsc{MQA} and \textsc{El-Att} would require 2/3, 1/3 and 1/3 of that number respectively. In contrast, \textsc{MHE} only accounts for 1\% of the \textsc{MHA} parameters.

\begin{table}[!t]
\begin{center}
\small
\setlength{\tabcolsep}{3pt}
\strutlongstacks{T}
\begin{tabular*}{\linewidth}{@{\extracolsep{\fill}\quad}lccc}
\toprule

\textbf{Attention} &\textbf{Complexity} & \textbf{\#Params} & \textbf{\#Params (+)} \\ \midrule

\enskip\enskip \textsc{SHA} & $\mathcal{O}(n)$ & $3d^2n$ & 0 \\
\enskip\enskip \textsc{MHA} & $\mathcal{O}(n^2)$ & $3d^2n^2$ & $(3n^2-3n)d^2$\\

\midrule

\enskip\enskip \textsc{EL-att} & $\mathcal{O}(n^2)$ & $d^2n^2$ & $(n^2-3n)d^2$\\
\enskip\enskip \textsc{MQA} & $\mathcal{O}(n^2)$ & $d^2n^2+2d^2n$ & $(n^2-n)d^2$\\
\enskip\enskip \textsc{SKV} & $\mathcal{O}(n^2)$ & $2d^2n^2$ & $(2n^2-3n)d^2$\\

\midrule

\multicolumn{3}{l}{\textsc{MHE} (ours)}\\
\enskip\enskip \textsc{-Add} & $\mathcal{O}(n)$ & $3d^2n+3dn$ & $3nd$\\
\enskip\enskip \textsc{-Mul} & $\mathcal{O}(n)$ & $3d^2n+3dn$ & $3nd$\\

\bottomrule
\end{tabular*}

\caption{Memory complexity regarding the number of parameters in each attention sublayer, while fixing the dimension of attention heads to $d$. $n$ denotes the number of attention heads. To simplify, the dimension of hidden states $d_m$ is set to $nd$. The last projection for pooling attention heads is excluded.} 

\label{table:save_result}
\end{center}
\end{table}

\begin{figure}[!t]
    \centering
    \includegraphics[width=0.97\linewidth]{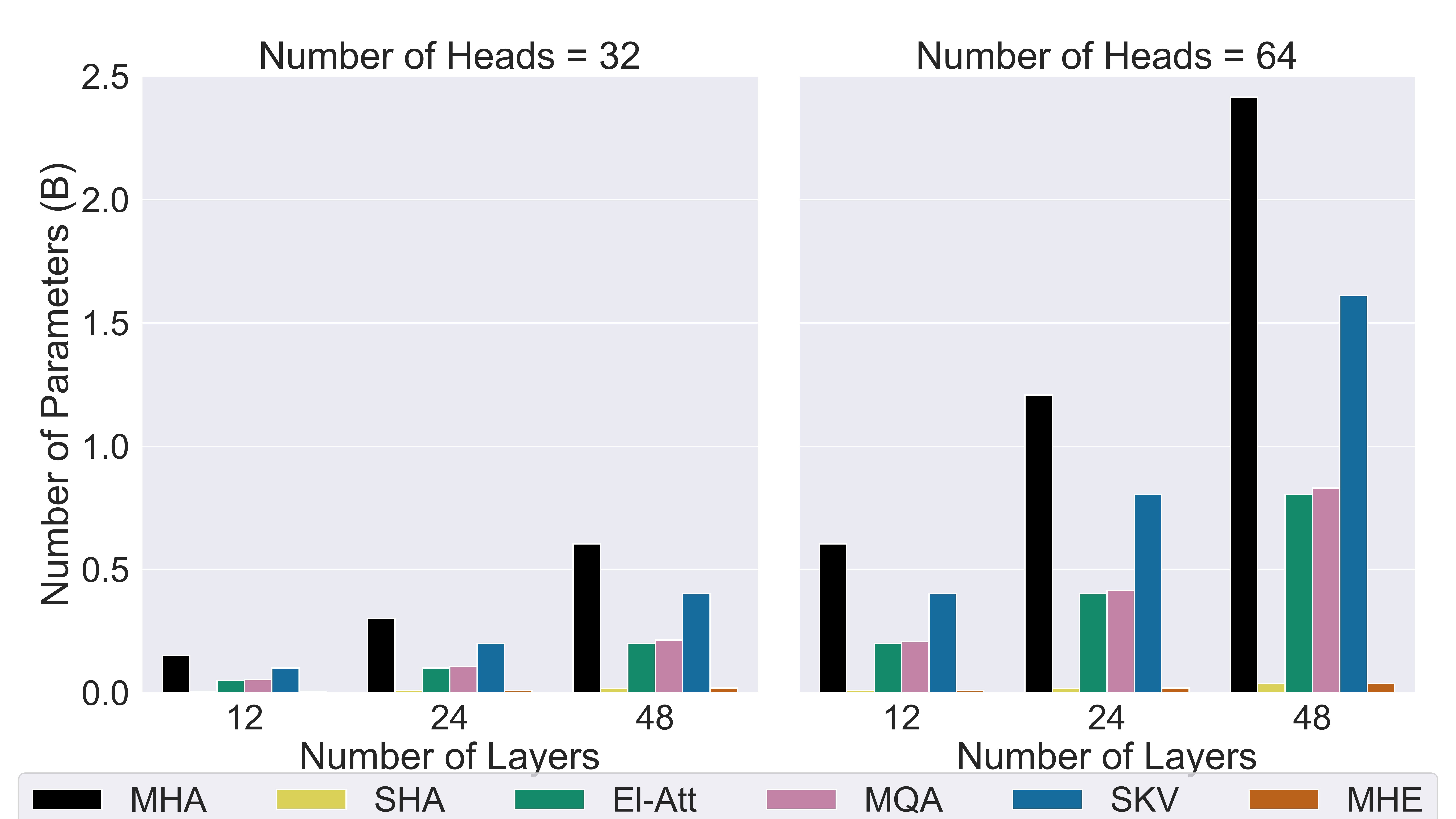}
    \caption{Total number of parameters in attention sublayers, while scaling the number of attention layers to 12, 24 and 48 with 32 attention heads and 64 attention heads respectively. We fix the dimension of attention to 64.
    }
    \label{fig:scaling_params_bar}
\end{figure}

Moreover, we also present in Figure \ref{fig:scaling_params_bar} the total number of parameters required across attention variants when stacking 12, 24 and 48 layers along with 32 and 64 attention heads respectively. We also fix the dimension of attention heads to 64. We can observe, when the number of attention head reaches 64, \textsc{MHA} with 24 layers already occupies more than 1B parameters, while \textsc{El-Att} and \textsc{MQA} reach 0.8B parameters with 48 layers. \textsc{SKV} takes 24 layers to reach 0.8B parameters. However, the total number of parameters in \textsc{MHE} attention does not exceed 0.1B even when scaling to 48 layers with 64 attention heads. It is also clear that scaling the attention module to 48 layers, 32 attention heads and 12 layers needs a comparable number of parameters for \textsc{MHA}, \textsc{El-Att}, \textsc{MQA} or \textsc{SKV}. This indicates, that LLM developers have to make a choice whether doubling the number of attention heads or cutting down the number of layers to a quarter when working under a tight memory budget. However, \textsc{MHE} does not suffer by such issues.

Further, we project these estimates to the popular GPT-3 model~\citep{brown2020language}. It is a decoder-only model with 96 decoder layers, 96 attention heads per layer, and a head dimension of 128. The vanilla multi-head attention module requires a massive 43.48B parameters. However,  using \textsc{MHE} attention, this number can be significantly reduced to 0.46B parameters, i.e. approximately a reduction by 98.9\%.\footnote{It would have been great to report results by pre-training our own \textsc{MHE} GPT-3 model, however this is prohibitive with the modest compute we have available.} Comparing this to other parameter-efficient attention variants such as \textsc{EL-att} (14.50B parameters), \textsc{MQA} attention (14.80B parameters), and \textsc{SKV} attention (28.99B parameters), it becomes evident that our \textsc{MHE} offers better memory efficiency. This makes it a compelling alternative for memory-constrained scenarios. See Appendix \ref{sec:scaling_robustness} for a detailed study on the robustness of MHE to model size changes (i.e. scaling).

\section{Discussion}

\textsc{MHA} enables the model to attend to information from different representation subspaces at different positions~\citep{vaswani2017attention}. It uses distinct projection matrices for each attention head and integrates the information from these different representation subspaces. However, \citet{vaswani2017attention} did not explore different methods for performing space transformations per head.

Previous work has pointed out that over-parameterized models might have a low intrinsic dimension. Therefore, transforming the projection matrices to smaller low-rank ones usually does not severely harm model predictive performance~\citep{li2018measuring, aghajanyan2020intrinsic}. Meanwhile, the classic \textsc{MHA} approach also does not impose any constraints on the orthogonality of these subspaces during pre-training and fine-tuning. The column vectors in those projection matrices could be highly collinear, i.e. the projection matrices could be rank-deficient. As a result, its inner-working mechanism could be simply understood as introducing levels of variation to the encoded representation of the same token at the same position across different heads.

Our \textsc{MHE} approach is possible to achieve memory efficiency (similar to \textsc{SHA}) together with high PRR compared to \textsc{MHA} by mimicking the position embeddings for representing different attention heads.

On one hand, the addition operation in \textsc{MHE-Add} is used for transforming the keys, queries and values. This can be seen as a small distortion of the subspace obtained through projection, followed by rotation. For an input representation, the difference between the projected and injected (i.e. through head embedding addition) queries, keys and values is a constant vector across any pair of heads. On the other hand, the \textsc{MHE-Mul} approach employs a multiplication operation, which more aggressively distorts and reshapes the keys, queries and values subspaces. Head embeddings in \textsc{MHE-Mul} play a role as the scaling factors, respectively stretching each dimension of the input representation. Thus, the difference between the keys, queries, and values generated by different heads for the same input representation, is a vector parallel to the projected input. This vector is dependent on the specific input, unlike the constant vector in \textsc{MHE-Add}.

Interestingly, our experimental results consistently show that the multiplication operation outperforms addition in the majority of benchmarks.
This corroborates findings of a previous empirical study by \citet{su2021roformer} that compared rotary position embeddings (somehow analogous to \textsc{MHE-Mul}) with absolute position embeddings (analogous to \textsc{MHE-Add}).

\section{Conclusions}

We have proposed \textsc{MHE} attention that employs a single shared projection matrix along with multiple head embeddings, to simplify and reduce the memory footprint of the \textsc{MHA}. Our experimental results have demonstrated that \textsc{MHE} attention exhibits superior memory efficiency compared to other memory-efficient attention variants,  while achieving high predictive performance ratio to \textsc{MHA} on various downstream tasks. Compared to a single-head attention, \textsc{MHA} requires $(3n^2-3n)d^2$ parameters for $n$ attention heads and head dimensionality $d$, while \textsc{MHE} barely requires a negligible $3nd$. For future research, we plan to investigate scaling up \textsc{MHE} models and explore its linguistic capabilities~\citep{vulic-etal-2020-probing,koto-etal-2021-discourse}.

\section*{Limitations}
We experiment only using `base' size models without experimenting with larger architectures, due to limited access to computational resources. Similarly, we did not experiment with decoder only architectures~\citep{brown2020language} which we leave for future work. We have not combined our \textsc{MHE} method with computationally efficient attention methods with linear complexity, such as Linformer~\citep{wang2020linformer}. We expect that it would speed up computation of \textsc{MHE}, but it is out of the scope of our paper.

\section*{Acknowledgments}
We would like to thank Constantinos Karouzos, Miles Williams and the anonymous reviewers for their invaluable feedback.

\bibliography{anthology,custom}
\bibliographystyle{acl_natbib}

\clearpage
\appendix

\section{Reported Metrics for Each Task}\label{sec:metrics}
We evaluate all models on \textsc{Glue}~\citep{wang-etal-2018-glue}, \textsc{SuperGlue}~\citep{wang2019superglue}, \textsc{SQuAD v1.1}~\citep{rajpurkar-etal-2016-squad} and \textsc{SQuAD v2.0}~\citep{rajpurkar-etal-2018-know}. We report matched accuracy for MNLI, Matthews correlation for CoLA, Spearman correlation for STS, F1 score for QQP, CB, MultiRC and \textsc{SQuAD} and accuracy for all other tasks. Table \ref{table:glue_result} and Table \ref{table:superglue_result} present results on \textsc{Glue} and \textsc{SuperGlue} respectively for our \textsc{MHE-Formers} models and all baselines with the encoder-only architecture. Table \ref{table:memory_result_tc_glue} and \ref{table:memory_result_tc_sg} present results of the scores and performance elasticity of parameters (PEoP) across all models over each task in \textsc{Glue} and \textsc{SuperGlue}. Table \ref{table:memory_result_tc_gpt} presents results on \textsc{Glue} for our \textsc{MHE-Formers} models and all baselines with the decoder-only architecture. Table \ref{table:memory_result_tc_gpt} presents results of the scores and performance elasticity of parameters (PEoP) across all models over each task in \textsc{Glue}.

\renewcommand*{\arraystretch}{1.1}
\begin{table*}[!t]
\begin{center}
\resizebox{\linewidth}{!}{
\strutlongstacks{T}
\begin{tabular}{l|rrrrrrrr|r}
\toprule
\textbf{ATTENTION} & \textbf{MNLI} & \textbf{QNLI} & \textbf{QQP} & \textbf{RTE} & \textbf{SST} & \textbf{MRPC} & \textbf{CoLA} & \textbf{STS} & \textbf{\textsc{Glue} Avg.} \\ \midrule
\textsc{SHA} & 80.5(0.3) & 87.5(0.2) & 86.7(0.1) & 63.6(0.9) & 90.7(0.3) & 85.1(0.7) & 53.8(1.1) & 85.8(0.4) & 79.2(0.1)\\
\textsc{MHA} & 83.4(0.1) & 89.8(0.3) & 87.8(0.1) & 67.6(1.5) & 92.0(0.3) & 86.8(0.4) & 59.6(1.3) & 88.5(0.3) & 81.9(0.3)\\\midrule
\textsc{EL-att} & 81.7(0.1) & 88.4(0.2)	& 87.3(0.2) & 67.6(1.0) & \textbf{91.7}(0.6) & 85.9(0.7) & 52.4(1.7) & 87.7(0.2) & 80.3(0.3)\\
\textsc{MQA} & \textbf{82.6}(0.1) & 88.8(0.2) & 87.3(0.1) & 66.5(0.9) & 91.4(0.5) & 87.3(0.2) & \textbf{58.4}(1.3) & 87.9(0.2) & 81.3(0.2)\\
\textsc{SKV} & \textbf{82.6}(0.1) & \textbf{89.4}(0.3) & \textbf{87.7}(0.1) & \textbf{68.2}(1.7) & 91.6(0.3) & \textbf{87.4}(0.6) & 56.2(1.2) & \textbf{88.6}(0.2) & \textbf{81.4}(0.2)\\ \hdashline
\textsc{FNet} & 76.3(0.1) & 83.8(0.1) & 84.8(0.1) & 63.2(2.0) & 88.4(0.7) & 78.0(0.4) & 43.2(2.5) & 83.7(0.3) & 75.2(0.6)\\
\textsc{Linear} & 75.4(0.1) & 81.4(0.1)	& 85.5(0.2)	& 54.7(2.3) & 90.4(0.4) & 72.2(0.6) & 50.3(1.0) & 70.9(0.5) & 72.6(1.1)\\
  \hdashline
\textsc{MHE-Add} & 81.5(0.2) & 87.8(0.2) & 87.2(0.1) & 66.9(2.0) & 90.5(0.4) & 87.2(0.3) & 54.7(0.9) & 87.7(0.1) & 80.4(0.2)\\
\textsc{MHE-Mul} & 81.9(0.1) & 87.9(0.1) & 87.4(0.1) & 67.1(1.5) & 91.1(0.5) & 85.4(0.5) & 56.6(1.7) & 87.3(0.2) & 80.6(0.2)\\

\midrule\midrule
\textsc{MHA}(M) & 84.4(0.2) & 91.1(0.4) & 84.0(0.6) & 70.5(1.0) & 92.0(0.2) & 87.2(0.8) & 62.5(1.0) & 88.8(0.2) & 82.6(0.4)\\
\textsc{MHE-Mul} (M) & 82.7(0.2) & 89.2(0.4) & 87.2(0.2) & 67.9(0.4) & 90.7(0.3) & 86.3(1.0) & 59.8(1.8) & 88.0(0.2) & 81.5(0.3)\\

\bottomrule
\end{tabular}
}
\caption{Results for encoder-only models on \textsc{Glue} dev sets with standard deviations over five runs in parentheses. \textbf{Bold} values denote best performing method in each task.} 

\label{table:glue_result}
\end{center}
\end{table*}

\renewcommand*{\arraystretch}{1.1}
\begin{table*}[!t]
\begin{center}
\resizebox{\linewidth}{!}{
\strutlongstacks{T}
\begin{tabular}{l|rrrrrrr|r}
\toprule
\textbf{ATTENTION} & \textbf{BoolQ} & \textbf{CB} & \textbf{RTE} & \textbf{WiC} & \textbf{MultiRC} & \textbf{COPA} & \textbf{WSC} & \textbf{\textsc{SuperGlue} Avg.} \\\midrule
\textsc{SHA} & 72.3(0.7) & 88.7(2.5) & 62.5(1.0) & 63.6(0.5) & 59.7(15.7) & 59.2(2.8) & 63.8(1.5) & 67.1(2.4)\\
\textsc{MHA} & 76.6(0.6) & 89.4(1.8) & 67.9(1.3) & 65.4(0.8) & 69.0(1.2) & 64.0(3.1) & 61.5(3.3) & 70.5(0.5)\\\midrule
\textsc{EL-att} & 73.5(1.0) & 85.7(4.8) & \textbf{69.5}(1.4) & 63.8(0.8) & 67.9(0.3) & 62.2(1.8) & 63.8(0.5) & 69.5(1.0)\\
\textsc{MQA} & 74.6(0.6) & 86.7(1.6) & 65.4(0.8) & 64.0(1.2) & \textbf{68.8}(0.5) & 62.2(2.0) & 63.3(2.7) & 69.3(0.7)\\
\textsc{SKV} & \textbf{75.2}(0.3) & 84.5(3.7) & 67.5(0.9) & \textbf{65.2}(1.1) & 68.7(0.2) & \textbf{64.0}(1.0) & \textbf{64.4}(1.2) & \textbf{69.9}(0.4)\\ \hdashline
\textsc{FNet} & 68.4(0.5) & 51.8(4.3) & 60.7(0.9) & 63.8(1.1) & 62.3(0.6) & 58.2(3.7) & 60.0(1.6) & 60.7(0.6)\\
\textsc{Linear} & 70.4(0.2)	& 50.6(2.1)	& 55.2(1.8)	& 62.9(0.7) & 57.8(0.5) & 60.0(2.8) & 61.0(1.1) & 59.7(0.9)\\ \hdashline
\textsc{MHE-Add} & 73.3(0.2) & 88.8(1.7) & 67.5(1.5) & 64.2(0.5) & 67.1(0.2) & 60.2(2.8) & 62.5(1.4) & 69.1(0.5)\\
\textsc{MHE-Mul} & 74.9(0.6) & \textbf{89.4}(1.0) & 67.8(1.3) & 64.7(0.6) & 68.0(0.3) & 61.6(1.5) & 61.2(2.9) & 69.6(0.3)\\

\midrule\midrule
\textsc{MHA}(M) & 78.1(0.3)	& 88.1(6.8) & 70.3(1.3) & 67.8(0.8) & 72.9(0.6) & 68.2(4.1) & 64.6(2.9) & 72.9(0.6)\\
\textsc{MHE-Mul} (M) & 75.2(0.5) & 84.6(2.4) & 68.6(1.8) & 66.3(0.9) & 69.8(0.4) & 61.6(3.8) & 64.6(0.8) & 70.1(1.1)\\

\bottomrule
\end{tabular}
}
\caption{Results for encoder-only models on \textsc{SuperGlue} dev sets with standard deviations over five runs in parentheses. \textbf{Bold} values denote best performing method in each task.} 

\label{table:superglue_result}
\end{center}
\end{table*}

\renewcommand*{\arraystretch}{1.1}
\begin{table*}[!t]
\begin{center}
\resizebox{\linewidth}{!}{
\strutlongstacks{T}
\begin{tabular}{l|llllllll}
\toprule

\textbf{ATTEN} & \multicolumn{8}{c}{\textsc{Glue}}\\
\textbf{-TION} & \textbf{MNLI} & \textbf{QNLI} & \textbf{QQP} & \textbf{RTE} & \textbf{SST} & \textbf{MRPC} & \textbf{CoLA}	& \textbf{STS} \\ \midrule
\textsc{SHA}  &  80.5\,- & 87.5\,- & 86.7\,- & 63.6\,- & 90.7\,- & 85.1\,- & 53.8\,- & 85.8\,-\\
\textsc{MHA}  &  83.4\,\,\,\,(0.02) & 89.8\,\,\,\,(0.01) & 87.8\,\,\,\,(0.01) & 67.6\,\,\,\,(0.03) & 92.0\,\,\,\,(0.01) & 86.8\,\,\,\,(0.01) & 59.6\,\,\,\,(0.05) & 88.5\,\,\,\,(0.01)\\\midrule
\textsc{EL-att} & 81.7\,\,\,\,(0.02) & 88.4\,\,\,\,(0.02) & 87.3\,\,\,\,(0.01) & 67.6\,\,\,\,(0.10) & \underline{91.7}\,\,\,\,(0.02) & 85.9\,\,\,\,(0.02) & 52.4\,\,(-0.04) & 87.7\,\,\,\,(0.04)\\
\textsc{MQA} & \underline{82.6}\,\,\,\,(0.04) & 88.8\,\,\,\,(0.02) & 87.3\,\,\,\,(0.01) & 66.5\,\,\,\,(0.06) & 91.4\,\,\,\,(0.01) & 87.3\,\,\,\,(0.04) & \underline{58.4}\,\,\,\,(0.12) & 87.9\,\,\,\,(0.03)\\
\textsc{SKV} & \underline{82.6}\,\,\,\,(0.02) & \underline{89.4}\,\,\,\,(0.02) & \underline{87.7}\,\,\,\,(0.01) & \underline{68.2}\,\,\,\,(0.05) & 91.6\,\,\,\,(0.01) & \underline{87.4}\,\,\,\,(0.02) & 56.2\,\,\,\,(0.03) & \underline{88.6}\,\,\,\,(0.02)\\ \hdashline

\textsc{FNet} & 76.3\,\,\,\,\,\,\,\,\,\,\,\,(-) & 83.8\,\,\,\,\,\,\,\,\,\,\,\,(-) & 84.8\,\,\,\,\,\,\,\,\,\,\,\,(-) & 63.2\,\,\,\,\,\,\,\,\,\,\,\,(-) & 88.4\,\,\,\,\,\,\,\,\,\,\,\,(-) & 78.0\,\,\,\,\,\,\,\,\,\,\,\,(-) & 43.2\,\,\,\,\,\,\,\,\,\,\,\,(-) & 83.7\,\,\,\,\,\,\,\,\,\,\,\,(-)\\

\textsc{Linear} & 75.4\,\,\,\,\,\,\,\,\,\,\,\,(-) & 81.4\,\,\,\,\,\,\,\,\,\,\,\,(-) & 85.5\,\,\,\,\,\,\,\,\,\,\,\,(-) & 54.7\,\,\,\,\,\,\,\,\,\,\,\,(-) & 90.4\,\,\,\,\,\,\,\,\,\,\,\,(-) & 72.2\,\,\,\,\,\,\,\,\,\,\,\,(-) & 50.3\,\,\,\,\,\,\,\,\,\,\,\,(-) & 70.9\,\,\,\,\,\,\,\,\,\,\,\,(-)\\ \hdashline

\textsc{MHE-Add} & 81.5\,\,\,\,(3.88) & 87.8\,\,\,\,(1.34) & 87.2\,\,\,\,(1.86) & 66.9\,(16.35) & 90.5\,\,(-0.81) & 87.2\,\,\,\,\textbf{(7.93)} & 54.7\,\,\,\,(5.22) & 87.7\,\,\,\,\textbf{(7.05)}\\
\textsc{MHE-Mul} & 81.9\,\,\,\,\textbf{(\textbf5.41)} & 87.9\,\,\,\,\textbf{(1.51)} & 87.4\,\,\,\,\textbf{(2.54)} & 67.1\,\textbf{(17.80)} & 91.1\,\,\,\,\textbf{(1.29)} & 85.4\,\,\,\,(1.29) & 56.6\,\textbf{(16.29)} & 87.3\,\,\,\,(5.60)\\

\bottomrule
\end{tabular}
}
\caption{Detailed average scores and performance elasticity of parameters (in parentheses) on \textsc{Glue} for \textsc{MHE} models and the baselines with encoder-only architecture using MLM as pre-training objectives. \textbf{Underlined} values denote the best performing method and \textbf{bold} values denote the method with best PEoP in each task.}

\label{table:memory_result_tc_glue}
\end{center}
\end{table*}

\renewcommand*{\arraystretch}{1.1}
\begin{table*}[!t]
\begin{center}
\resizebox{\linewidth}{!}{
\strutlongstacks{T}
\begin{tabular}{l|lllllll}
\toprule

\textbf{ATTEN} & \multicolumn{7}{c}{SuperGlue}\\
\textbf{-TION} & \textbf{BoolQ} & \textbf{CB} & \textbf{RTE} & \textbf{WIC} & \textbf{MultiRC} & \textbf{COPA} & \textbf{WSC} \\ \midrule
\textsc{SHA}  &  72.3\,- & 88.7\,- & 62.5\,- & 63.6\,- & 59.7\,- & 59.2\,- & 63.8\,-\\
\textsc{MHA}  &  76.6\,\,\,\,(0.03) & 89.4\,\,\,\,(0.00) & 67.9\,\,\,\,(0.04) & 65.4\,\,\,\,(0.01) & 69.0\,\,\,\,(0.07) & 64.0\,\,\,\,(0.04) & 61.5\,(-0.02)\\\midrule
\textsc{EL-att} & 73.5\,\,\,\,(0.03) & 85.7\,\,(-0.06) & \underline{69.5}\,\,\,\,(0.19) & 63.8\,\,\,\,(0.00) & 67.9\,\,\,\,(0.23) & 62.2\,\,\,\,(0.08) & 63.8\,\,\,\,(0.00)\\
\textsc{MQA} & 74.6\,\,\,\,(0.04) & 86.7\,\,(-0.03) & 65.4\,\,\,\,(0.06) & 64.0\,\,\,\,(0.01) & \underline{68.8}\,\,\,\,(0.21) & 62.2\,\,\,\,(0.07) & 63.3\,(-0.01) \\
\textsc{SKV} & \underline{75.2}\,\,\,\,(0.03) & 84.5\,\,(-0.03) & 67.5\,\,\,\,(0.06) & \underline{65.2}\,\,\,\,(0.02) & 68.7\,\,\,\,(0.11) & \underline{64.0}\,\,\,\,(0.06) & \underline{64.4}\,\,\,\,\textbf{(0.01)} \\ \hdashline

\textsc{FNet} & 68.4\,\,\,\,\,\,\,\,\,\,\,\,(-) & 51.8\,\,\,\,\,\,\,\,\,\,\,\,(-) & 60.7\,\,\,\,\,\,\,\,\,\,\,\,(-) & 63.8\,\,\,\,\,\,\,\,\,\,\,\,(-) & 62.3\,\,\,\,\,\,\,\,\,\,\,\,(-) & 58.2\,\,\,\,\,\,\,\,\,\,\,\,(-) & 60.0\,\,\,\,\,\,\,\,\,\,\,\,(-) \\

\textsc{Linear} & 70.4\,\,\,\,\,\,\,\,\,\,\,\,(-) & 50.6\,\,\,\,\,\,\,\,\,\,\,\,(-) & 55.2\,\,\,\,\,\,\,\,\,\,\,\,(-) & 62.9\,\,\,\,\,\,\,\,\,\,\,\,(-) & 57.8\,\,\,\,\,\,\,\,\,\,\,\,(-) & 60.0\,\,\,\,\,\,\,\,\,\,\,\,(-) & 61.0\,\,\,\,\,\,\,\,\,\,\,\,(-) \\ \hdashline

\textsc{MHE-Add} & 73.3\,\,\,\,(4.58) & 88.8\,\,\,\,(0.54) & 67.5\,(25.50) & 64.2\,\,\,\,(3.00) & 67.1\,(39.90) & 60.2\,\,\,\,(5.41) & 62.5\,\,(-6.75) \\
\textsc{MHE-Mul} & 74.9\,\textbf{(11.78)} & \underline{89.4}\,\,\,\,\textbf{(2.52)} & 67.8\,\textbf{(26.97)} & 64.7\,\,\,\,\textbf{(5.36)} & 68.0\,\textbf{(44.63)} & 61.6\,\textbf{(12.97)} & 61.2(-13.49)\\

\bottomrule
\end{tabular}
}
\caption{Detailed average scores and performance elasticity of parameters (in parentheses) on \textsc{SuperGlue} for \textsc{MHE} models and the baselines with encoder-only architecture using MLM as pre-training objectives. \textbf{Underlined} values denote the best performing method and \textbf{bold} values denote the method with best PEoP in each task.}

\label{table:memory_result_tc_sg}
\end{center}
\end{table*}

\renewcommand*{\arraystretch}{1.1}
\begin{table*}[!t]
\begin{center}
\resizebox{\linewidth}{!}{
\strutlongstacks{T}
\begin{tabular}{l|rrrrrrrr|r}
\toprule
\textbf{ATTENTION} & \textbf{MNLI} & \textbf{QNLI} & \textbf{QQP} & \textbf{RTE} & \textbf{SST} & \textbf{MRPC} & \textbf{CoLA} & \textbf{STS} & \textbf{\textsc{Glue} Avg.} \\ \midrule
\textsc{SHA} & 78.7(0.1) & 86.0(0.2) & 85.0(0.1) & 66.5(0.9) & 89.8(0.2) & 76.8(0.4) & 38.0(1.3) & 81.5(0.4) & 75.3(0.3)\\
\textsc{MHA} & 80.6(0.1) & 87.9(0.2) & 86.3(0.1) & 66.9(1.1) & 90.2(0.3) & 79.0(0.7) & 42.9(1.3) & 86.0(0.2) & 77.5(0.2)\\\midrule
\textsc{EL-att} & 79.5(0.2)	& 86.8(0.3)	& 85.7(0.1)	& 65.7(1.4)	& 90.0(0.4) & 79.2(1.4) & 41.5(2.2) & 84.3(0.2) & 76.6(0.4)\\
\textsc{MQA} & 80.0(0.1) & 86.3(0.1) & \textbf{85.9}(0.1) & 66.2(0.7) & 90.3(0.3) & \textbf{80.7}(0.6) & 41.3(0.8) & 84.3(0.4) & 76.9(0.2)\\
\textsc{SKV} & \textbf{80.3}(0.1) & \textbf{87.5}(0.3) & \textbf{85.9}(0.1) & 66.1(1.1) & 90.6(0.5) & 79.6(0.5) & \textbf{41.9}(1.8) & \textbf{84.9}(0.2) & \textbf{77.1}(0.4)\\ \hdashline
\textsc{MHE-Add} & 78.7(0.1) & 85.6(0.2) & 85.4(0.1) & 66.4(2.5) & 89.6(0.4) & 78.7(0.6) & 38.4(1.2) & 83.5(0.3) & 75.8(0.3)\\
\textsc{MHE-Mul} & 79.0(0.2) & 85.5(0.1) & 85.6(0.1) & \textbf{70.2}(2.5) & \textbf{90.9}(0.2) & 78.9(0.8) & 39.4(1.3) & 84.0(0.3) & 76.7(0.2)\\

\bottomrule
\end{tabular}
}
\caption{Results for decoder-only models on \textsc{Glue} dev sets with standard deviations over five runs in parentheses. \textbf{Bold} values denote best performing method in each task.} 

\label{table:glue_result_gpt}
\end{center}
\end{table*}

\renewcommand*{\arraystretch}{1.1}
\begin{table*}[!t]
\small
\begin{center}
\resizebox{\linewidth}{!}{
\strutlongstacks{T}
\begin{tabular}{l|llllllll}
\toprule

\textbf{ATTEN} & \multicolumn{8}{c}{\textsc{Glue}}\\
\textbf{-TION} & \textbf{MNLI} & \textbf{QNLI} & \textbf{QQP} & \textbf{RTE} & \textbf{SST} & \textbf{MRPC} & \textbf{CoLA}	& \textbf{STS}\\ \midrule
\textsc{SHA} & 78.7\,- & 86.0\,- & 85.0\,- & 66.5\,- & 89.8\,- & 76.8\,- & 38.0\,- & 81.5\,-\\
\textsc{MHA} & 80.6\,\,\,\,(0.01) & 87.9\,\,\,\,(0.01) & 86.3\,\,\,\,(0.01) & 66.9\,\,\,\,(0.00) & 90.2\,\,\,\,(0.00) & 79.0\,\,\,\,(0.01) & 42.9\,\,\,\,(0.06) & 86.0\,\,\,\,(0.03)\\\midrule
\textsc{EL-att} & 79.5\,\,\,\,(0.02) & 86.8\,\,\,\,\textbf{(0.02)} & 85.7\,\,\,\,(0.01) & 65.7\,\,(-0.02) & 90.0\,\,\,\,(0.01) & 79.2\,\,\,\,(0.05) & 41.5\,\,\,\,(0.16) & 84.3\,\,\,\,(0.06)\\
\textsc{MQA} & 80.0\,\,\,\,(0.02) & 86.3\,\,\,\,(0.01) & \underline{85.9}\,\,\,\,(0.01) & 66.2\,\,(-0.01) & 90.3\,\,\,\,(0.01) & \underline{80.7}\,\,\,\,(0.07) & 41.3\,\,\,\,(0.12) & 84.3\,\,\,\,(0.05)\\
\textsc{SKV} & \underline{80.3}\,\,\,\,(0.01) & \underline{87.5}\,\,\,\,(0.01) & \underline{85.9}\,\,\,\,(0.01) & 66.1\,\,(-0.00) & 90.6\,\,\,\,(0.01) & 79.6\,\,\,\,(0.03) & \underline{41.9}\,\,\,\,(0.07) & \underline{84.9}\,\,\,\,(0.03)\\ \hdashline

\textsc{MHE-Add} & 78.7\,\,(-0.02) & 85.6\,\,(-1.50) & 85.4\,\,\,\,(1.60) & 66.4\,\,(-0.69) & 89.6\,\,(-0.57) & 78.7\,\,\,\,(7.96) & 38.4\,\,\,\,(3.71) & 83.5\,\,\,\,(7.98)\\
\textsc{MHE-Mul} & 79.0\,\,\,\,\textbf{(0.98)} & 85.5\,\,(-1.88) & 85.6\,\,\,\,\textbf{(2.05)} & \underline{70.2}\,\textbf{(17.72)} & \underline{90.9}\,\,\,\,\textbf{(4.01)} & 78.9\,\,\,\,\textbf{(8.58)} & 39.4\,\textbf{(12.32)} & 84.0\,\,\,\,\textbf{(9.97)}\\

\bottomrule
\end{tabular}
}
\caption{Detailed average scores and performance elasticity of parameters (in parentheses) on \textsc{Glue} for \textsc{MHE} models and the baselines with decoder-only architecture using MLM as pre-training objectives. \textbf{Underlined} values denote the best performing method and \textbf{bold} values denote the method with best PEoP in each task.}

\label{table:memory_result_tc_gpt}
\end{center}
\end{table*}

\section{Hyperparameters}\label{sec:hyperparameters}
The hyperparameters used in pre-training are listed in Table \ref{table:hyperparams_pretraining}. The hyperparameters used in fine-tuning are listed in Table \ref{table:hyperparams_finetuning}.

\renewcommand*{\arraystretch}{1.0}
\begin{table}[!t]
\begin{center}
\small
\begin{tabular}{ll}
\toprule
Hyperparameter & Pretraining
\\ \midrule
Maximum train steps & 1000000 steps \\
Batch size (per GPU) & 32 instances \\
Adam $\epsilon$ & 1e-8 \\
Adam $\beta_1$ & 0.9 \\
Adam $\beta_2$ & 0.9999 \\
Sequence length & 512 \\
Peak learning rate & 1e-4 for MLM \\
Learning rate schedule & linear \\
Warmup steps & 10000 \\
Weight decay & 0.01 \\
Attention Dropout & 0.1 \\
Dropout & 0.1 \\

\bottomrule
\end{tabular}
\caption{Details of hyperparameters used in pre-training.} 

\label{table:hyperparams_pretraining}
\end{center}
\end{table}

\renewcommand*{\arraystretch}{1.0}
\begin{table}[!t]
\begin{center}
\small
\resizebox{\linewidth}{!}{
\begin{tabular}{ll}
\toprule
Hyperparameter & Fine-tuning
\\ \midrule
Maximum train epochs & \makecell[l]{20 epochs for \textsc{Glue}, \textsc{SuperGlue} and \textsc{SQuAD}}\\
Batch size (per GPU) & 32 instances \\
Adam $\epsilon$ & 1e-6 \\
Adam $\beta_1$ & 0.9 \\
Adam $\beta_2$ & 0.999 \\
Peak learning rate & \makecell[l]{3e-5 for \textsc{Glue} and \textsc{SQuAD};\\
5e-5 for \textsc{SuperGlue}}\\
Learning rate schedule & cosine with hard restarts \\
Warmup steps & \makecell[l]{first 6\% steps for \textsc{Glue} and \textsc{SuperGlue};\\3327 for \textsc{SQuAD v1.1};\\4950 for \textsc{SQuAD v2.0} 
}\\
Weight decay & 0 \\
Attention Dropout & 0.1 \\
Dropout & 0.1 \\
Evaluation steps & 2455 for MNLI, 655 for QNLI, \\
& 2275 for QQP, 48 for RTE, \\
& 421 for SST, 69 for MRPC, \\
& 162 for CoLA and 108 for STS,\\
& 177 for BoolQ, 5 for CB, \\
& 47 for RTE, 102 for WiC, \\
& 512 for MultiRC, 8 for COPA, \\
& 11 for WSC,\\
& 548 for \textsc{SQuAD v1.1}, \\
& 815 for \textsc{SQuAD v2.0} \\

\bottomrule
\end{tabular}
}
\caption{Details of hyperparameters used in fine-tuning.} 

\label{table:hyperparams_finetuning}
\end{center}
\end{table}

\section{Memory Usage}\label{sec:memory_usage}
To further illustrate the memory-efficiency of our \textsc{MHE} models compared to the baselines, we take the \textsc{BERT}-base architecture (12 attention heads, each with a dimension of 64) as an example, and measure the memory usage per attention block as in Section 2.1.1 from \citet{smith2022using} and report the memory usage saving ratio (\%) during the attention calculation in Table \ref{table:memory_usage}:

\renewcommand*{\arraystretch}{1.1}
\begin{table*}[!t]
\begin{center}
\resizebox{\linewidth}{!}{
\strutlongstacks{T}
\begin{tabular}{l|rrrr|rr}
\toprule
\textbf{ATTENTION} & \textbf{weights} & \textbf{gradients} & \textbf{Adam states} & \textbf{activations} & \textbf{Total} & \textbf{Memory Saving Ratios} (\%) \\ \midrule
\textsc{SHA} & 4423680 & 4423680 & 5898240 & 25165824 & 39911424 & 44.84\\
\textsc{MHA} & 14155776 & 14155776 & 18874368 & 25165824 & 72351744 & 0.00\\\midrule
\textsc{EL-att} & 7077888 & 7077888 & 9437184 & 25165824 & 48758784 & 32.61\\
\textsc{MQA} & 7667712 & 7667712 & 10223616 & 25165824 & 50724864 & 29.89\\
\textsc{SKV} & 10616832 & 10616832 & 14155776 & 25165824 & 60555264 & 16.30\\ \hdashline
\textsc{MHE-Add} & 4437504 & 4437504 & 5916672 & 25165824 & 39957504 & 44.77\\
\textsc{MHE-Mul} & 4437504 & 4437504 & 5916672 & 25165824 & 39957504 & 44.77\\

\bottomrule
\end{tabular}
}
\caption{Memory usage (in bytes) and memory saving ratios (compared to \textsc{MHA}) per attention block for our \textsc{MHE} and other baselines. \textsc{MHA} denotes \textsc{BERT}-base here.}

\label{table:memory_usage}
\end{center}
\end{table*}

The calculation is based on inputs with batch size of 32, hidden dimension of 768, sequence length of 512 and fp16 mixture precision training using the following formula: 
\begin{itemize}
\item Memory(weights)=\#params*(2+4) bytes;
\item Memory(gradients)=\#params*(2+4) bytes; 
\item Memory(Adam states)=\#params*(4+4) bytes;
\item Memory(activations)= batch-size*sequence-length*hidden-dimension*2 bytes.
\end{itemize}

From Table \ref{table:memory_usage}, we observe the memory usage saving ratio of our proposed \textsc{MHE} is 2.75 times better than \textsc{SKV}, 1.50 times better than \textsc{MQA} and 1.37 times better than \textsc{El-Att}, which indicates a SotA memory saving capabilities compared to all other parameter-efficient attention variants.

\section{Robustness to Scaling}\label{sec:scaling_robustness}
We also conduct experiments to observe the effectiveness and the robustness of our best \textsc{MHE-Mul} while scaling the model size.

Table \ref{table:size_scaling} presents average accuracy on two text classification benchmarks (\textsc{Glue} and \textsc{SuperGlue}), perplexities on two language modelling benchmarks (\textsc{WikiText-103} and \textsc{Penn Treebank}) with their corresponding performance retention ratio (PRR) for \textsc{MHA} and \textsc{MHE-MUl} in both encoder-only and decoder-only architecture across different model sizes.\footnote{\textsc{BASE}: 12 encoder/decoder layers, each containing 12 attention heads; \textsc{LARGE}/\textsc{MEDIUM}: 24 encoder/decoder layers, each containing 16 attention heads.} For the encoder-only models, we observe that the PRR of \textsc{MHE-Mul} remains stable on \textsc{Glue} (from 98.4\% to 98.7\%) and \textsc{SuperGlue} (from 98.7\% to 96.2\%) while scaling the number of parameters in the attention  blocks to 3.5 times larger. For the decoder-only models, the PRR on \textsc{Glue} for \textsc{MHE-MUl} stabilizes at 97.9\% (i.e. 1.1\% lower) after scaling. Surprisingly, the PRR of \textsc{MHE-Mul} increases on \textsc{WikiText-103} (from 74.9\% to 95.2\%) and \textsc{Penn Treebank} (from 85.6\% to 88.5\%) while scaling to \textsc{MEDIUM} size.

Similar results are observed for the encoder-decoder architecture on \textsc{WMT14} machine translation task. According to Table \ref{table:seq2seq_scaling}, we first observe the PRR of \textsc{MHE-Mul} remains stable (i.e. between 91.5\% and 96.0\%) across all different sizes, where the number of parameters in the corresponding \textsc{MHA} ranges from 19.87M to 75.50M. Meanwhile, we also observe that making the model deeper by stacking more encoder and decoder layers results in a steady increment on PRR for \textsc{MHE-Mul} (e.g. 93.6\%, 95.0\% and 96.0\% respectively, for 8 layers, 12 layers and 16 layers in total). Moreover, for the same number of parameters in the attention, wider attention heads consistently leads to a better PRR for \textsc{MHE-Mul}, i.e. 91.5\%, 95.0\% and 95.3\% for the dimensionality of 32, 64 and 128 of attention heads respectively.

These results indicate \textsc{MHE} consistently achieves good performance retention ratios and is robust to model size change.

\begin{table*}[!t]
\begin{center}

\resizebox{\linewidth}{!}{
\strutlongstacks{T}
\begin{tabular}{ll|rr|rrr|rrr|rrr|rrr}
\toprule
 & & \multicolumn{2}{c|}{\textbf{\#Params}(M)} & \multicolumn{3}{c|}{\textbf{\textsc{Glue}}} & \multicolumn{3}{c|}{\textbf{\textsc{SuperGlue}}} & \multicolumn{3}{c|}{\textbf{\textsc{WikiText-103}}} & \multicolumn{3}{c}{\textbf{\textsc{Penn Treebank}}}\\

 & & \makecell[c]{\textsc{MHA}} & \multicolumn{1}{c|}{\makecell[c]{\textsc{MHE}\\\textsc{-Mul}}} & \makecell[c]{\textsc{MHA}} & \makecell[c]{\makecell[c]{\textsc{MHE}\\\textsc{-Mul}}} & \multicolumn{1}{c|}{\textbf{PRR}} & \makecell[c]{\textsc{MHA}} & \makecell[c]{\makecell[c]{\textsc{MHE}\\\textsc{-Mul}}} & \multicolumn{1}{c|}{\textbf{PRR}} & \makecell[c]{\textsc{MHA}} & \makecell[c]{\makecell[c]{\textsc{MHE}\\\textsc{-Mul}}} & \multicolumn{1}{c|}{\textbf{PRR}} & \makecell[c]{\textsc{MHA}} & \makecell[c]{\makecell[c]{\textsc{MHE}\\\textsc{-Mul}}} & \makecell[c]{\textbf{PRR}}\\\midrule

Encoder & \textsc{base} & 28.32 & 8.88 & 81.9 & 80.6 & 98.4 & 70.5 & 69.6 & 98.7 & - & - & - & - & - & -\\
-only & \textsc{large} & 100.66 & 29.96 & 81.5 & 82.6 & 98.7 & 72.9 & 70.1 & 96.2 & - & - & - & - & - & -\\\midrule
Decoder & \textsc{base} & 28.32 & 8.88 & 77.5 & 76.7 & 99.0 & - & - & - & 43.0 & 53.8 & 74.9 & 44.3 & 50.7 & 85.6\\
-only & \textsc{medium} & 100.66 & 29.96 & 79.4 & 77.7 & 97.9 & - & - & - & 35.5 & 37.2 & 95.2 & 37.5 & 41.6 & 88.5\\

\bottomrule
\end{tabular}
}
\caption{Results of evaluation metrics on two text classification benchmarks (\textsc{Glue}, \textsc{SuperGlue}) and two language modelling benchmarks (\textsc{WikiText-103} and \textsc{Penn Treebank}) with performance retention ratio (PRR) for \textsc{MHA} and \textsc{MHE-MUl} across different model sizes.} 

\label{table:size_scaling}
\end{center}
\end{table*}

\begin{table*}[!t]
\begin{center}
\small

\resizebox{\linewidth}{!}{
\strutlongstacks{T}
\begin{tabular}{l|rrrrlc|rr|rr|r}
\toprule

 & \makecell[c]{$\mathbf{N}$} & \makecell[c]{$\mathbf{d_m}$} & \makecell[c]{$\mathbf{h}$} & \makecell[c]{$\mathbf{d_h}$} & \makecell[c]{$\mathbf{p_{drop}}$} & \makecell[c]{\textbf{\#Steps}} & \multicolumn{2}{c|}{\textbf{\#Params}(M)} 
 & \multicolumn{2}{c|}{\textbf{BLEU}} & \makecell[c]{\textbf{PRR}}\\
 & \multicolumn{6}{c|}{ } & \makecell[c]{\textsc{MHA}} & \makecell[c]{\textsc{MHE-Mul}} & \makecell[c]{\textsc{MHA}} & \makecell[c]{\textsc{MHE-Mul}} &\\\midrule

\textsc{base} & 12 & 512 & 8	& 64 & 0.1 & 100K & 18.87 & 6.52 & 24.8 & 23.6 & 95.0\\
& 12	& 512 & 16 & 32	& 0.1 & 100K & 18.87 & 5.63	& 25.1 & 22.9 & 91.5\\
& 12	& 512 & 4 & 128	& 0.1 & 100K & 18.87 & 8.29	& 24.7 & 23.6 & 95.3\\\hdashline
\textsc{4L} & 8 & 512 & 8 & 64 & 0.1 & 100K & 12.58 & 4.34 & 23.9 & 22.4 & 93.6\\
\textsc{8L} & 16	& 512 & 8 & 64 & 0.1 & 100K	& 25.17	& 8.69 & 25.3 & 24.3 & 96.0\\\hdashline
\textsc{12H} & 12 & 768 & 12	& 64 & 0.15	& 100K & 42.47 & 13.31 & 25.7 & 24.2 & 94.2\\\hdashline
\textsc{big} & 12 & 1024	& 16 & 64 & 0.3	& 300K & 75.50 & 22.47 & 26.5 & 24.8 & 93.6\\\bottomrule
\end{tabular}
}
\caption{Results of BLEU scores on \textsc{WMT-14} English to German machine translation task with performance retention ratio (PRR) for \textsc{MHA} and \textsc{MHE-MUl} across different model sizes.} 

\label{table:seq2seq_scaling}
\end{center}
\end{table*}

\end{document}